\documentclass[letterpaper]{article} 
\usepackage{aaai23}  
\usepackage{times}  
\usepackage{helvet}  
\usepackage{courier}  
\usepackage[hyphens]{url}  
\usepackage{graphicx} 
\usepackage{natbib}  
\usepackage{caption} 
\usepackage{algorithm}
\usepackage{algorithmic}

\usepackage{subfig}
\usepackage{booktabs}
\usepackage{multirow}
\usepackage{multicol}
\usepackage{arydshln}

\usepackage{newfloat}
\usepackage{listings}
\DeclareCaptionStyle{ruled}{labelfont=normalfont,labelsep=colon,strut=off} 
\floatstyle{ruled}
\newfloat{listing}{tb}{lst}{}
\floatname{listing}{Listing}
\title{Learning from Training Dynamics: Identifying Mislabeled Data \\ Beyond Manually Designed Features}
\author{
    Qingrui Jia\textsuperscript{\rm 1,\rm 2\footnote{The first two authors contributed equally to this work.}}, 
    Xuhong Li\textsuperscript{\rm 2\footnotemark[1]},
    Lei Yu\textsuperscript{\rm 1,\rm 3},
    Jiang Bian\textsuperscript{\rm 2},
    Penghao Zhao\textsuperscript{\rm 2},
    Shupeng Li\textsuperscript{\rm 2},\\
    Haoyi Xiong\textsuperscript{\rm 2\footnote{Corresponding author.}},
    Dejing Dou\textsuperscript{\rm 4}
}
\affiliations{
    \textsuperscript{\rm 1}
    Sino-French Engineer School,
    Beihang University,
    Beijing, China~
    \textsuperscript{\rm 2}
    Baidu Inc.
    Beijing, China\\
    \textsuperscript{\rm 3}
    Beihang Hangzhou Innovation Institute Yuhang,
    Hangzhou, China~
    \textsuperscript{\rm 4}
    BCG Greater China,
    Beijing, China\\
    
   \{jiaqingrui, lixuhong, xionghaoyi\}@baidu.com, 
   \{jiaqr, yulei\}@buaa.edu.cn
}

\usepackage{bibentry}

\usepackage{amsmath,amsfonts,bm}

\def\eqref#1{equation~\ref{#1}}

\def\1{\bm{1}}

\def\vf{{\bm{f}}}
\def\vg{{\bm{g}}}

\def\mL{{\bm{L}}}

\DeclareMathAlphabet{\mathsfit}{\encodingdefault}{\sfdefault}{m}{sl}
\SetMathAlphabet{\mathsfit}{bold}{\encodingdefault}{\sfdefault}{bx}{n}

\def\gD{{\mathcal{D}}}

\def\gL{{\mathcal{L}}}

\DeclareMathOperator*{\argmin}{arg\,min}

\begin{document}

\maketitle

\begin{abstract}
While mislabeled or ambiguously-labeled samples in the training set could negatively affect the performance of deep models, diagnosing the dataset and identifying mislabeled samples helps to improve the generalization power. 
\emph{Training dynamics}, i.e., the traces left by iterations of optimization algorithms, have recently been proved to be effective to localize mislabeled samples with hand-crafted features.
In this paper, beyond manually designed features, we introduce a novel learning-based solution, leveraging a \emph{noise detector}, instanced by an LSTM network, which learns to predict whether a sample was mislabeled using the raw training dynamics as input. 
Specifically, the proposed method trains the noise detector in a supervised manner using the dataset with synthesized label noises and can adapt to various datasets (either naturally or synthesized label-noised) without retraining. 
We conduct extensive experiments to evaluate the proposed method.
We train the noise detector based on the synthesized label-noised CIFAR dataset and test such noise detector on Tiny ImageNet, CUB-200, Caltech-256, WebVision and Clothing1M. 
Results show that the proposed method precisely detects mislabeled samples on various datasets without further adaptation, and outperforms state-of-the-art methods.
Besides, more experiments demonstrate that the mislabel identification can guide a label correction, namely data debugging, providing orthogonal improvements of algorithm-centric state-of-the-art techniques from the data aspect.



\end{abstract}

\section{Introduction}

Deep learning models have achieved remarkable performances across various tasks~\cite{lecun2015deep} due to the rapid development of paralleled devices, advanced network architectures and learning algorithms, as well as the large dataset collection and high-quality annotations~\cite{deng2009imagenet,DBLP:journals/corr/webvision-abs-1708-02862,DBLP:journals/pami/Places-ZhouLKO018}.
However, the difficulty and expensiveness of acquiring high-quality labels inevitably leave an amount of label noises on the datasets.
In many real-world settings, training samples and their labels are collected through proxy variables or web scraping~\cite{DBLP:conf/iccv/ChenG15,DBLP:conf/cvpr/Clothing-XiaoXYHW15,DBLP:conf/eccv/JoulinMJV16,DBLP:journals/corr/webvision-abs-1708-02862}, which certainly leads to a larger amount of noises.
According to recent studies~\citep{zhang2021understanding,DBLP:conf/icml/ArpitJBKBKMFCBL17,DBLP:conf/icml/AroraDHLW19}, datasets with poor-quality labels, including mislabeled and ambiguously labeled training samples, would negatively affect the generalization performance of deep models.

\begin{figure*}[t]
\centering
\includegraphics[width=1.95\columnwidth]{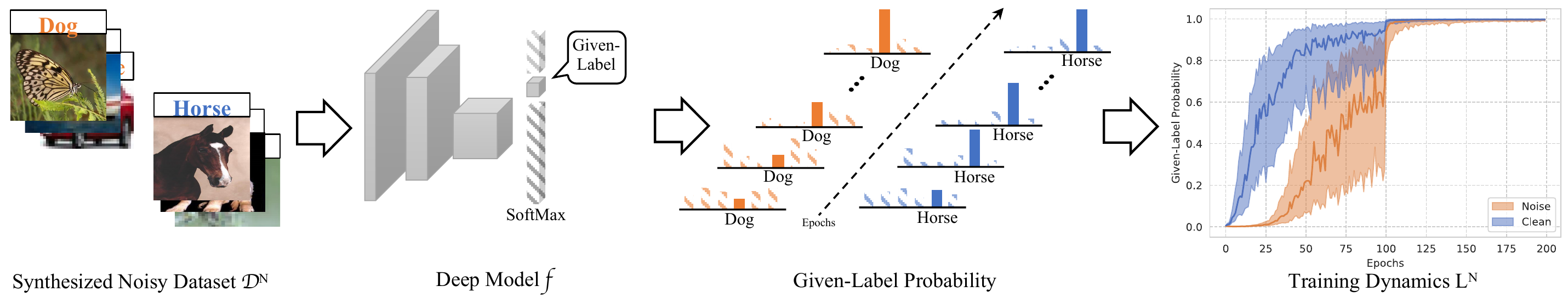}
\caption{Acquisition of training dynamics. The plots on the right are averaged across 1000 random samples from the synthesized-noise Caltech-256 dataset, with the training epoch index as X-axis and the probability of the given-label class as Y-axis.}
\label{fig:illustration}
\vspace{-1em}
\end{figure*}

To address the label noises, two lines of researches have been proposed.
One is from the \textit{algorithm} aspect~\citep{song2022learning}, including robust objective functions~\citep{DBLP:conf/iclr/PereyraTCKH17,DBLP:conf/aaai/GhoshKS17,DBLP:conf/iccv/0001MCLY019}, sample reweight~\citep{DBLP:conf/cvpr/PatriniRMNQ17,DBLP:conf/icml/MentorNet} and hybrid approaches~\citep{DBLP:conf/iclr/DivideMix-LiSH20,DBLP:conf/iclr/SELF-NguyenMNNBB20,DBLP:conf/cvpr/NishiDRH21,DBLP:conf/iclr/XiaL00WGC21}.
Another line is from the \textit{data} aspect, focusing on identifying mislabeled samples and improving the quality of datasets.
A series of methods following this line are based on the observation that in the context of Stochastic Gradient Descent (SGD), hand-crafted features on \textbf{training dynamics} are able to separate the mislabeled samples from clean ones~\citep{toneva2018empirical,pleiss2020identifying,DBLP:conf/emnlp/Cartography-SwayamdiptaSLWH20,DBLP:conf/nips/PaulGD21}.
Subtle differences on the training dynamics can be discerned, as in Figure~\ref{fig:illustration}, where the averaged probability of being classified into the labeled category (given-label probability) across mislabeled samples is lower and their variance is higher compared to the clean ones.
More distinguishable features of training dynamics can even characterize the learning difficulty of each training sample~\citep{DBLP:conf/icml/HacohenCW20,DBLP:conf/nips/BaldockMN21}.

However, the aforementioned distinguishable features are usually extracted manually from the training dynamics with prior knowledge and domain expertise, and it is not always fitting well and being efficient.
Tracing back the history of machine learning, the deep learning methods are much more effective than traditional learning with handcraft features, by learning features from raw data and solving the learning tasks in end-to-end fashions.
In light of the evolution of machine learning, we propose a novel solution through \emph{learning from training dynamics} to identify mislabeled samples.
Beyond manually designed features, our proposed method\footnote{Code available at https://github.com/Christophe-Jia/mislabel-detection and at InterpretDL~\citep{JMLR:v23:21-0738} as well.} trains a \emph{noise detector}, instanced by an LSTM network~\citep{DBLP:journals/neco/LSTM-HochreiterS97}, on the raw training dynamics, with the objective of learning distinguishable features to solve the binary classification task, i.e., separating mislabeled samples from clean ones.

The most efficient and effective way to train the noise detector is the supervised learning, where the binary annotations whether samples being mislabeled or not are required.
However, this kind of information is usually not provided by public datasets.
To cope with this issue, we purposely wrong-set the labels of certain samples from an (almost) clean dataset, and conduct the standard training process on it.
In this way, the raw training dynamics can be trained with the supervision of the synthesized label noises.

We carry out three types of evaluation experiments to validate the effectiveness of our method in identifying mislabeled samples.
The results show that (1) The noise detector trained on the synthesized noisy CIFAR-100 can precisely identify the mislabels on other synthesized and real-world noisy datasets without further adaptation or retraining, showing a good transfer ability; (2) Compared to the state-of-the-art approaches from the data aspect, our method outperforms with a clear margin; and (3) Combining with methods from the algorithm aspect, our method provides an orthogonal improvement over the current methods.

In summary, our main contributions are the following:
\begin{itemize}
    \vspace*{-0.2em}
    \setlength\itemsep{-0.1em}
    \item To identify the mislabeled samples, we propose a novel supervised-learning-based solution to train an LSTM-instanced label \emph{noise detector} based on the synthesized noisy dataset and its training dynamics.
    
    \item The noise detector trained on a noised CIFAR-100 is an off-the-shelf detector for identifying the mislabeled samples, i.e., it can accurately find label noises without any further adaptation or retraining on other noised datasets, including Tiny ImageNet, CUB-200, Caltech-256, WebVision and Clothing 100K, where the latter two are real-world noisy datasets.
    
    \item On three types of evaluation experiments, our proposed method outperforms all the existing methods of identifying the mislabeled samples. Analyses and discussions are also provided to validate the effectiveness of our method.
    
    \item Combing with the learning-with-noisy-labels algorithms, our proposed method further boosts test accuracies of the models trained on noisy datasets. This also proves that the improvements from the data aspect can be orthogonal to those from the algorithm aspect.
    
\end{itemize}

\section{Related Work}

Learning with label noise has achieved remarkable success. We review a good amount of prior works most relevant to ours and categorize them into two aspects, \textit{algorithms} or \textit{data}. For a complete review, we refer to~\citet{song2022learning,DBLP:journals/kbs/AlganU21,mazumder2022dataperf}.

\textbf{From Algorithms Aspect.} Tremendous approaches provide algorithms-centric perspectives and stay in the conventional mainstream for robust learning in presence of mislabeled samples.
Some approaches proposed to estimate the noise transition matrix via pretrained model \cite{DBLP:conf/cvpr/PatriniRMNQ17}, clean validation set \cite{DBLP:conf/nips/HendrycksMWG18} or expectation-maximization (EM) algorithm \cite{DBLP:conf/iclr/GoldbergerB17}. 
A number of methods propose to improve the noise immunity by designing robust loss functions~\citep{DBLP:conf/aaai/GhoshKS17,DBLP:conf/iccv/0001MCLY019}, or modify loss or probabilities~\citep{DBLP:conf/icml/Loss-Correction-ArazoOAOM19,yao2019safeguarded,reed2014training} to compensate for negative impact owing to noisy labels.
Some alternative methodologies aim to refurbish the corrupted labels, by characterizing the distribution of noisy labels~\citep{DBLP:conf/cvpr/Clothing-XiaoXYHW15,vahdat2017toward,li2017learning,DBLP:conf/cvpr/CleanNet} or in a meta-learning manner~\cite{DBLP:conf/iccv/LiYSCLL17,DBLP:conf/aaai/ZhengAD21}.

Remarkably, the state-of-the-art methods are usually hybrid, integrated of above components.
For examples, DivideMix~\citep{DBLP:conf/iclr/DivideMix-LiSH20} leverages Gaussian mixture model to distinguish mislabel ones, then regards them as unlabeled samples, and trains the model in a semi-supervised learning manner. 
\citet{DBLP:conf/cvpr/NishiDRH21} improves the DivideMix framework by applying different data augmentation strategies for finding noisy labels and learning models.
Through curriculum learning~\citep{bengio2009curriculum}, RoCL~\citep{DBLP:conf/iclr/ZhouWB21} 
first learns from clean data and then gradually involves noisy-labeled data corrected by the predictions from an ensemble based model.

In brief, algorithm-centric training with label noises requires a modification of the optimization algorithm.
The proposed approach in this paper, from a data-centric aspect, does not modify the optimization procedure but filters the label noises in the first step.
By all means, algorithm-centric methods are used to improve the optimization procedures, orthogonal to data-centric methods introduced as follows.

\textbf{From Data Aspect. } Fundamentally different from the above methods in learning with label noise, a lot of approaches focus on data-centric amelioration and construct higher quality datasets, decoupled from training procedures.
A straightforward thought of harnessing mislabeled samples is excluding them from the training data and then retraining on the remaining samples~\citep{DBLP:journals/jair/BrodleyF99}.
Therefore, various studies endeavor to detect mislabeled samples or sort clean samples. 
One of the prevalent methods is unsupervised outliers removal~\citep{Liu_2014_CVPR,xia2015learning}, which is based on the assumption that outliers are incorrectly labeled. 

Recently, instead of simply checking the final predictions, more works analyze the \textit{training dynamics}, i.e., the traces left by the steps during SGD.
For instance, \citet{toneva2018empirical} defines a forgetting event using training dynamics, and samples prone to be forgotten with high frequency probably correspond to be mislabeled. \citet{DBLP:conf/emnlp/Cartography-SwayamdiptaSLWH20} builds a data map with training dynamics, then marks off a region composed of instances that the model finds hard to learn in dataset mapping, which might be labeling errors. \citet{pleiss2020identifying} 
proffers a statistic called area under margin (AUM) and introduces threshold samples to separate clean and mislabeled samples.
In addition, other solutions filter noisy data from perspective of influence functions~\citep{koh2017understanding} for identifying training points that exert the most influence on the model. \citet{DBLP:conf/icml/GhorbaniZ19} introduced data Shapley to quantify the value of each training data to the predictor performance and considered the low Shapley value data as corruptions.
All of them utilize statistics or manually designed features to diagnose the dataset.
In this paper, based on training dynamics, we propose a novel learning-based solution, that trains a noise detector on the synthesized noisy dataset and detects the mislabeled samples in the original dataset.


\section{The Proposed Approach}


Training dynamics are distinguishable features to identify the mislabeled samples.
As shown in Figure~\ref{fig:illustration}, the mean of the given-label probability over noisy samples is lower than the one over clean samples; but the variance is higher, especially at the latter half of the training, indicating that the mislabeled samples are more complicated to fit than the clean samples.
However, with these visually visible differences, it is still challenging to identify the mislabeled samples precisely.
Designing more expert features will surely be helpful, but instead, we propose to benefit from the advantage of deep learning models to learn good features for classifying the mislabeled samples from clean ones.

\begin{figure}[t]
\centering
\includegraphics[width=\columnwidth]{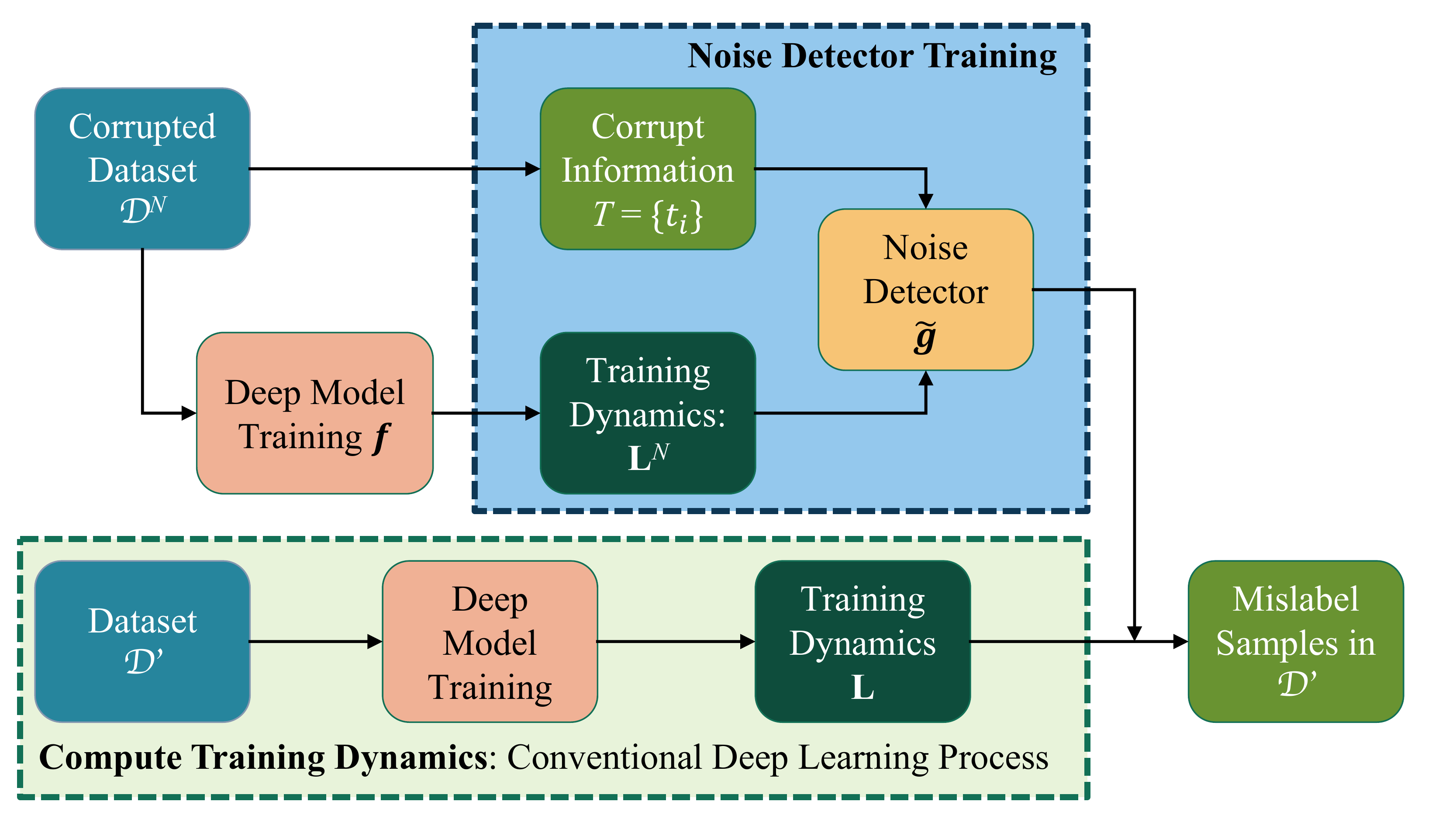}
\caption{Supervised Learning from Training Dynamics.}
\label{fig:pipeline}
\vspace{-1em}
\end{figure}


\subsection{Definition of Training Dynamics}

We give notations and definitions within the context of classification learning problems without loss of generality.

In the context of SGD or its variant, we define the \textbf{training dynamics} as the probability values of all training samples obtained by the model at each training epoch, denoted as $\mL$.
Specifically, $\mL(i, t)$ is the probabilities of the $i$-th sample at the $t$-th training epoch.
At each epoch, the model would pass through forward and backward propagation for each training sample, with the probability values as intermediate results that can be recorded during the training process without additional computations\footnote{For cases when the value of a certain sample at some epoch is missing, interpolations across training epochs can be used.}.
Note that $\mL(i, t)$ is a vector of $C$ elements, where $C$ is the number of classes, and we note $\mL_c(i, t)$ as the probability of the $c$-th class.
Noting the deep model as $\vf$, then we have 
\begin{equation}
    \mL(i, t) = \vf^{(t)}(x_i),
\end{equation}
where $x_i$ is the input data.
For computing gradients and updating the model's parameters, the label $y_i$ is needed.

\subsection{Supervised Learning from Training Dynamics}
\label{sec:method-steps}

We present the supervised learning from training dynamics in  three steps as following.

\textbf{Step 1.}
Given an (almost) clean dataset $\gD$, we first generate the label noises and obtain the noised dataset $\gD^N$.
Specifically, $\gD$ contains the data instances $(x_i, y_i)$ while $\gD^N$ contains $(x_i, y^N_i, k_i)$.
The additional term $k_i$ indicates that the sample is intentionally mislabeled ($k_i=1$) or not ($k_i=0$).
The design of the noise generator follows the previous works~\citep{pleiss2020identifying,DBLP:journals/jair/ConfidentLearning-NorthcuttJC21}, where the symmetric and asymmetric noises are used in our experiments.
Following the noise generator, the new values of $y^N_i$ will overwrite the original information in $\gD$, and $k_i$ can be simply obtained by
\begin{equation}
  k_i=  \begin{cases}
    1, & \text{if $y^N_i \neq y_i$};\\
    0, & \text{if $y^N_i = y_i$}.
  \end{cases}
\end{equation}

The noised datasets $\gD^N$ are only used to train the LSTM-instanced noise detector, denoted as $\vg$.
The trained noise detector $\tilde{\vg}$ (as introduced in Eq (\ref{eq:g})) is directly applicable to other real-world datasets.

As for the (almost) clean dataset $\gD$, though some famous datasets, such as CIFAR-10 and ImageNet, contain a small ratio ($<3\%$) of mislabeled samples~\citep{DBLP:conf/nips/NorthcuttAM21}, our method is well-adapted to a mild quantify of underlying noises in the dataset.
This is shown in Subsection~\ref{subsec:analyses}.

\textbf{Step 2.}
We train the deep model $\vf$ on $\gD^N$ using SGD or one of its variants, and record the training dynamics $\mL^N$.
We supervise the noise detector $\vg$ taking the training dynamics $\mL^N$ as feature inputs and the label noise information $K=\{k_i\}$ as supervision labels.
Specifically, the noise detector is optimized following
\begin{equation}
    \label{eq:g}
    \tilde{\vg} = \argmin_{\vg} \ \sum_i {\gL(\vg(\mL^N(i)), k_i)}\ ,
\end{equation}
where $\gL(\cdot,\cdot)$ is the binary cross entropy loss function, and the optimization process is guided by an AdamW optimizer~\citep{DBLP:journals/iclr/Adam-KingmaB14,DBLP:conf/iclr/AdamW-LoshchilovH19}.
Moreover, to deal with the epoch-wise training dynamics, $\vg$ is instanced by an LSTM model which is a popular choice for time-series sequences.
In practice, for efficiency and generality, we only take the given-label probability as input, i.e., $\mL_{y_i}^N(i)$.
Involving other classes would increase the parameters and computations in the LSTM model and the efficiency would be encumbered for datasets with over hundreds of classes.

\textbf{Step 3.}
Finally, we conduct the standard training process on the clean dataset $\gD'$, and record the training dynamics $\mL$.
The previously trained $\tilde{\vg}$ is then used to estimate the probability of the label $y_i$ being mislabeled according to the training dynamics $\mL_{y_i}(i)$, i.e., computing $\tilde{\vg}(\mL_{y_i}(i))$.

\begin{algorithm}[h]
    \caption{Supervised Learning from Training Dynamics.}
    \label{algo:main}
    \begin{algorithmic}[1]
       \STATE \textbf{Input:} $\gD$ an (almost) clean dataset for noise detector training, $\gD'$ datasets that requires mislabel identification, $\vf$ deep model architecture.
       
       \STATE \textbf{Step 1:} 
            \begin{ALC@g}
                \STATE Generate label noises based on $\gD$, and obtain $\gD^N$ for training the noise detector $\vg$ in Step 2.
            \end{ALC@g}
       
       \STATE \textbf{Step 2:} 
            \begin{ALC@g}
                \STATE Train $\vf$ on $\gD^N$, and get $\mL^N$.
                \STATE Train the noise detector $\vg$ using $\mL^N$ and $K=\{k_i\}$ according to Eq (\ref{eq:g}), and obtain $\tilde{\vg}$.
            \end{ALC@g}
       
       \STATE \textbf{Step 3:} 
            \begin{ALC@g}
                \STATE Train $\vf$ on $\gD'$ and get $\mL$.
                \STATE Compute $\tilde{\vg}(\mL_{y_i}(i))$ for each sample $(x_i, y_i)$ in $\gD'$, and predict whether $(x_i, y_i)$ is mislabeled.
            \end{ALC@g}

       \STATE \textbf{Output:} Predicted mislabeled samples in $\gD'$.
       
    \end{algorithmic}
\end{algorithm}

The procedure of the proposed approach is summarized in Algorithm~\ref{algo:main}.
We introduce three extensions of this proposed approach in the following subsection to show the applicability and robustness of our approach for practical usages.

\subsection{Characteristic Analysis}

The proposed approach effectively identifies the mislabeled samples, demonstrated by the experiments in Section~\ref{sec:exp}.
Here we provide the characteristic analysis to show the transferability, robustness and applicability of our method.




\textbf{Off-The-Shelf Noise Detector and Transferability.}
A trained $\tilde{\vg}$ on one dataset is capable of identifying the mislabeled samples on other datasets.
Once the noise detector $\tilde{\vg}$ is trained, \textbf{Step 3} can process standalone for any dataset $\gD'$, which proves that $\tilde{\vg}$ is generalizable and has learned some common features for identifying mislabeled samples.
In Section~\ref{sec:exp}, we show that $\tilde{\vg}$ trained from the noised CIFAR-100 works quite well on other (supposedly-) clean datasets and noisy datasets.
Note that further fine-tuning of $\tilde{\vg}$ on $\gD'$, following the exact procedure described in Section~\ref{sec:method-steps}, can always provide further improvements. 

\textbf{Robustness against Underlying Noises.}
In real-world scenarios, the dataset $\gD$ may contain a small or mild ratio of noises.
This would introduce noises in the training process of $\vg$.
We simulate this scenario by manually changing the labels of the (supposedly-)clean dataset and setting it to $\gD$.
Then we process the three steps described previously and use $\tilde{\vg}$ to detect the mislabeled samples.
According to our analytical experiments, the noise detector $\tilde{\vg}$ is still effective in this setting, up to a large ratio of noises.
This relaxes the constraint on the label quality of $\gD$.

\textbf{Usability of Super-Classes Noises.}
During the dataset annotation, human experts usually mislabel the samples into similar categories but rarely to those with very different categories.
This pattern of mislabeling is easy to mimic, especially in CIFAR-100, which has 20 super-classes, followed by 5 finer categories.
We follow this setting in Section~\ref{sec:method-steps} to generate the noisy labels within the super-class, instead of across all 100 classes.
Experiments show that our method is still capable of accurately identifying the mislabeled samples under this super-classes noise setting.


\section{Experiments}
\label{sec:exp}


On synthesized and real-world noisy datasets, three types of evaluations are conducted and presented in Section~\ref{subsec:identify}, \ref{subsec:real-world} and \ref{subsec:com with SSL}, respectively:
(1) On synthesized datasets, where the mislabeled samples are known, the precision and recall can be directly calculated.
(2) Following the procedure of previous works~\citep{DBLP:journals/jair/BrodleyF99,pleiss2020identifying}, the evaluations are performed by checking the performance of the supervised learning with/without (by excluding) mislabeled samples in the datasets. 
A higher test accuracy trained on the dataset without mislabeled samples indicates a better identification of mislabeled samples.
(3) Similar to the data debugging evaluation scheme~\citep{mazumder2022dataperf}, the evaluation procedure first identifies the label errors and then corrects them with the true labels or pseudo labels generated by the model in (2), i.e., trained excluding label errors. 
Afterward, the corrected dataset is incorporated with the state-of-the-art learning algorithms, e.g., DivideMix~\citep{DBLP:conf/iclr/DivideMix-LiSH20} and DM-AugDesc~\citep{DBLP:conf/cvpr/NishiDRH21}, to enjoy the orthogonal improvements from the data aspect. 



\begin{table*}[t]
\caption{Testing accuracy on the CIFAR-10/100 with symmetric label noises (ResNet-32).}
\label{tab:synthesized datasets}
\centering
\scalebox{1}{
    \begin{tabular}{lcccccccc}
    \toprule
    Dataset  & \multicolumn{4}{c}{CIFAR-10} & \multicolumn{4}{c}{CIFAR-100} \\
    Noise   & 0.2   & 0.4   & 0.6   & 0.8   & 0.2   & 0.4   & 0.6   & 0.8   \\ 
        \midrule 
        \midrule
    Standard & 75.0$\pm$0.3 & 56.7$\pm$0.4 & 36.7$\pm$0.4 & 16.6$\pm$0.2 & 49.6$\pm$0.2 & 37.5$\pm$0.2 & 23.8$\pm$0.4 & 8.2$\pm$0.3 \\ 
    DY-Bootstrap & 79.4$\pm$0.1 & 68.8$\pm$1.4 & 56.4$\pm$1.7 & Diverged & 53.0$\pm$0.4 & 43.0$\pm$0.3 & 36.6$\pm$0.5 & 12.8$\pm$0.5 \\
    Data Param & 82.1$\pm$0.2 & 70.8$\pm$0.9 & 49.3$\pm$0.7 & 18.9$\pm$0.3 & 56.3$\pm$1.4 & 46.1$\pm$0.4 & 32.8$\pm$2.3 & 11.9$\pm$1.2 \\
    INCV & 89.5$\pm$0.1 & 86.8$\pm$0.1 & 81.1$\pm$0.3 & 53.3$\pm$1.9 & 58.6$\pm$0.5 & 55.4$\pm$0.2 & 43.7$\pm$0.3 & 23.7$\pm$0.6 \\
    AUM & 90.2$\pm$0.0  & 87.5$\pm$0.1 & 82.1$\pm$0.0 & 54.4$\pm$1.6 & \textbf{65.5$\pm$0.2} & \textbf{61.3$\pm$0.1} & 53.0$\pm$0.5 & \textbf{31.7$\pm$0.7} \\
    Ours & \textbf{91.1$\pm$0.0} & \textbf{88.9$\pm$0.0} & \textbf{83.5$\pm$0.2} & \textbf{55.5$\pm$1.4} & \textbf{65.6$\pm$0.3} & 60.8$\pm$0.2 & \textbf{54.3$\pm$0.4} & 30.1$\pm0.6$ \\ 
    \midrule
    Oracle   & 91.0$\pm$0.1  & 90.3$\pm$0.4  & 89.2$\pm$0.9 & 87.4$\pm$1.7 & 64.5$\pm$0.1 & 61.0$\pm$0.2 & 55.2$\pm$0.5 & 44.9$\pm$0.4 \\ 
    \bottomrule
    \end{tabular}
}
\vspace{-1em}
\end{table*}

\subsection{Identify Mislabeled Samples}
\label{subsec:identify}
This section validates our approach on the datasets with synthesized label noises. We follow the commonly used noise distributions~\citep{DBLP:journals/jair/ConfidentLearning-NorthcuttJC21,pleiss2020identifying} to generate label noises at different ratios and change the labels of the training set with symmetry (or asymmetry) noises but leave the testing set unchanged.

Here we mainly compare our method with \textbf{AUM}~\citep{pleiss2020identifying}, one of the most effective approaches that also use the training dynamics to identify mislabeled samples.

\textbf{Experiment Setups.} 
Evaluations are conducted on four datasets with synthesized noisy labels: CIFAR-10/100~\citep{krizhevsky2009cifar}, CUB-200-2011~\cite{wah2011caltech} and Caltech256~\cite{griffin2007caltech}.
Evaluation metrics include \textbf{mAP}, \textbf{ROC AUC} scores and \textbf{Precision@95}, where \textbf{Precision@95} represents the precision when the recall reach 95\%, which can be directly measured because of the available ground-truth.
Training dynamics are insensitive to the deep model architectures~\citep{toneva2018empirical,pleiss2020identifying}.
We therefore choose ResNets as the deep model $\vf$ for the higher testing accuracy in original datasets, while we have also tested VGGs~\citep{simonyan2014very}, which show similar results in preliminary experiments.
Concerning the noise detector, for efficiency, an LSTM of two hidden layers with 64 dimensions for each is used.
All the noise detectors share the same architecture, allowing the possible transfer between datasets. 

\begin{figure}[t]
\centering
\includegraphics[width=0.98\columnwidth]{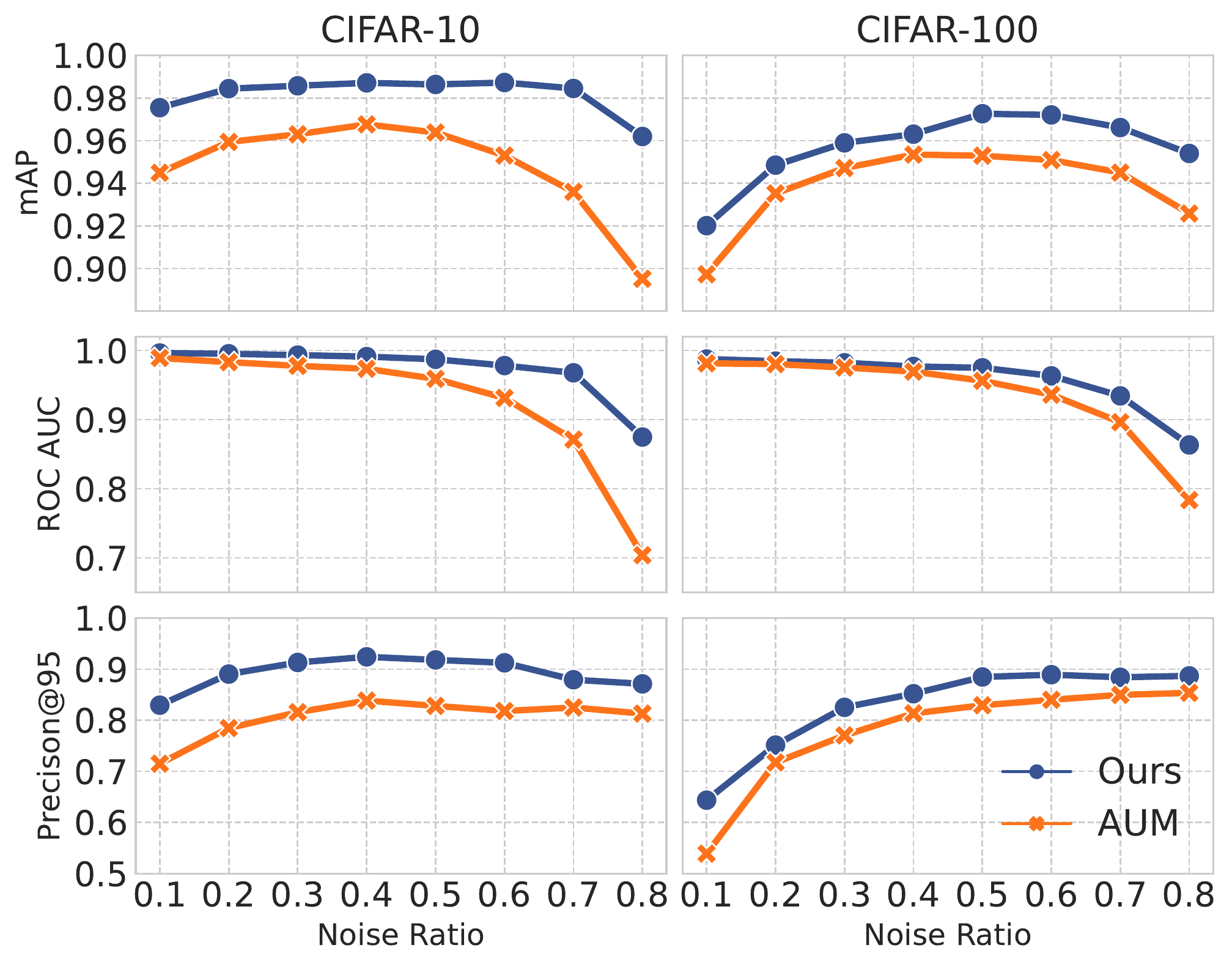}
\caption{Scores of mAP, ROC AUC and Precision@95 for identifying mislabeled samples on CIFAR-10/100 with symmetric label noises.}
\vspace{-1em}
\label{fig:map_auc_prec_cifar}
\end{figure}

\textbf{Results.}
Following the steps in Section~\ref{sec:method-steps}, we supervise the noise detector instanced by an LSTM model on the noised CIFAR-10/100 datasets with the corrupted labels.
Our method is competitive with AUM without threshold strategy.
Figure~\ref{fig:map_auc_prec_cifar} depicts the mAP, AUC and Precision@95 scores for identifying mislabeled samples from the noised CIFAR-10/100 datasets. 
Our procedure excels AUM in both symmetric and asymmetric noise settings, the later is confirmed in Figure~\ref{fig:asym}.

\begin{figure}[t]
\centering
\includegraphics[width=0.98\columnwidth]{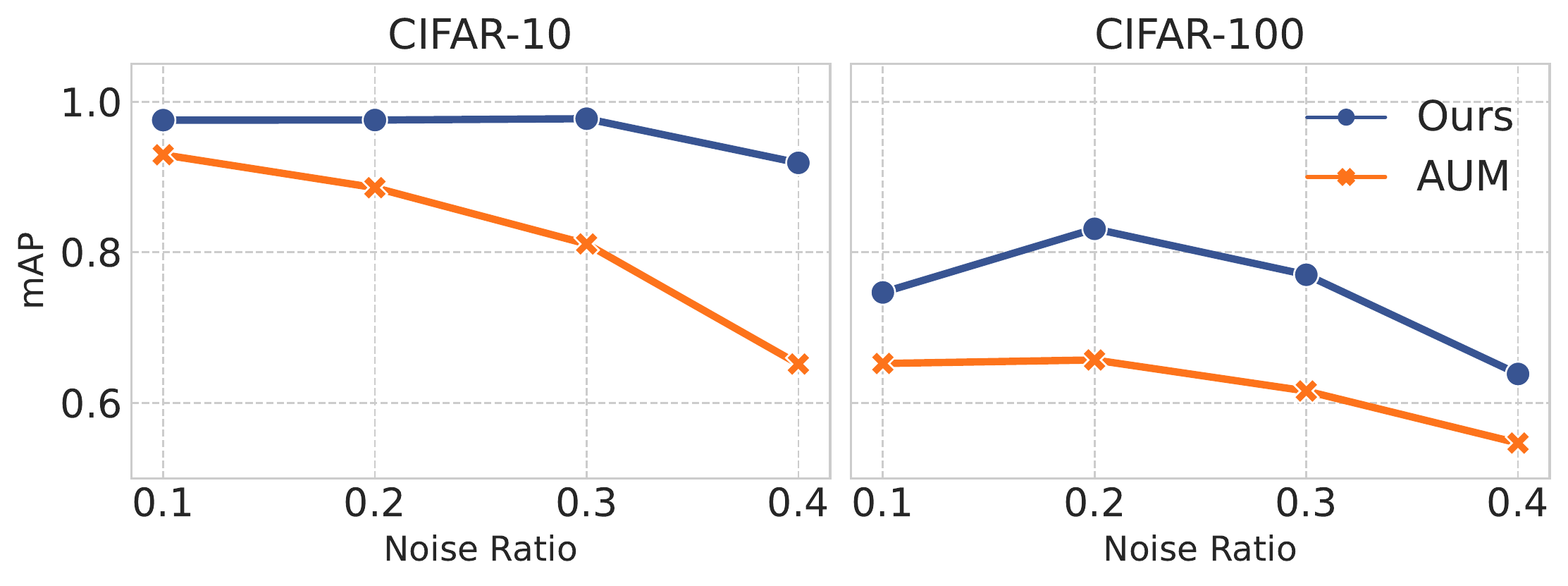}
\caption{Scores of mAP for identifying mislabeled samples on CIFAR-10/100 with asymmetric label noises.}
\vspace{-1em}
\label{fig:asym}
\end{figure}



\begin{figure}[t]
\centering
\includegraphics[width=0.98\columnwidth]{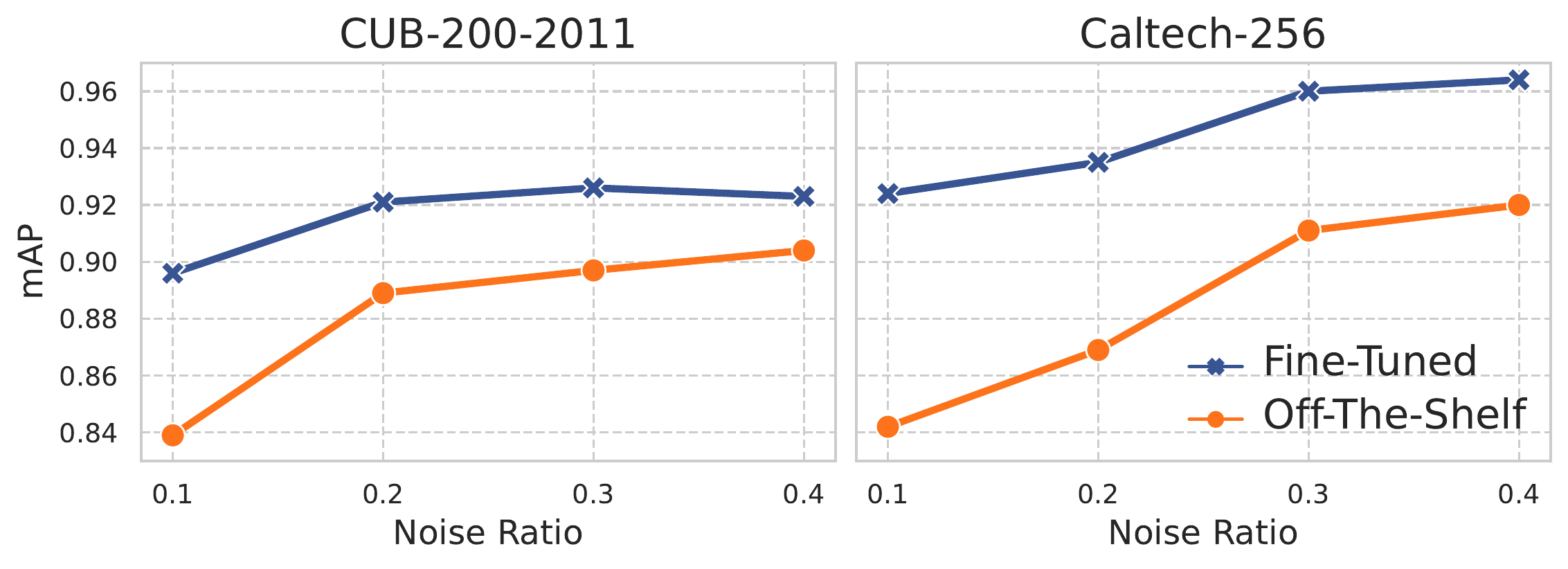}
\caption{Scores of mAP for fine-tuned and off-the-shelf detectors on two synthesized noisy CUB-200-2011 and Caltech-256.}
\vspace{-1em}
\label{fig:mAP migrated versus fine-tuning}
\end{figure}

Additional experiments have been conducted on the noised CUB-200-2011 and Caltech-256.
As shown in Figure~\ref{fig:mAP migrated versus fine-tuning}, our approach benefits from the well-trained noise detector $\tilde{\vg}$ on CIFAR-100 with 30\% noise ratio.
Then we fine-tune $\tilde{\vg}$ on the noised CUB-200-2011 and Caltech-256, respectively, to further improve the performance of the noise detection with a clear margin.


\begin{table*}[t]
    \caption{Testing accuracy on real-world datasets (ResNet-32 for CIFAR/Tiny I.N., ResNet-50 for others). }
	\label{tab:real-world datasets}
	\centering
    \scalebox{1.0}{
    \begin{tabular}{cccccc}
    \toprule
                 & CIFAR-10     & CIFAR-100    & Tiny ImageNet & WebVision50 & Clothing 100K \\ 
    \midrule
    \midrule    
    Standard     & 91.9$\pm$0.1  & 67.0$\pm$0.3 & 50.7$\pm$0.1  & 78.6         & 64.2        \\
    Data Param   & 91.9$\pm$0.0  & 63.6$\pm$1.4 & 51.6$\pm$0.2  & 78.5         & 64.5        \\
    DY-Bootstrap & 90.0$\pm$0.0 & 65.1$\pm$0.1 & 48.4$\pm$0.0  & 74.2         & 61.6        \\
    INCV         & 90.9$\pm$0.0  & 61.8$\pm$0.1 & 43.9$\pm$0.1  & 77.9         & 66.7        \\
    AUM          & 92.1$\pm$0.0  & 68.2$\pm$0.1 & 51.4$\pm$0.1  & 80.2         & 66.5        \\
    Ours & \textbf{92.5$\pm$0.1} & \textbf{68.7$\pm$0.3} & \textbf{52.2$\pm$0.0} & \textbf{81.4} & \textbf{68.2} \\ 
    \bottomrule
    \end{tabular}
    }
\vspace{-1em}
\end{table*}

\subsection{Retrain after Excluding Mislabeled Samples}
\label{subsec:real-world}

Mislabeled samples are harmful to the model~\citep{DBLP:conf/icml/AroraDHLW19,zhang2021understanding}, and removing them should be beneficial for its accuracy and generalizability.
To validate the effectiveness of identification, we remove samples predicted to be mislabeled using the proposed approach and then retrain the deep model. 
Based on the testing accuracy after excluding samples, we are able to measure the effectiveness of mislabeled sample identification.
Within this setting, we compare our method against several existing ones, introduced in the following paragraph.

\textbf{Baselines.} We compare our approach with several methods from the existing literature. 
\textbf{DY-Bootstrap}
~\citep{DBLP:conf/icml/Loss-Correction-ArazoOAOM19} proposes to fit a beta mixture model to estimate the probability that a sample is mislabeled.
\textbf{Data Param}~\citep{saxena2019data} allows a learnable parameter to each sample or class, which controls their importance during the training stage. 
\textbf{INCV}~\citep{DBLP:conf/icml/INCV-ChenLCZ19} presents a strategy to select a clean subset successively harnessing iterative cross-validation. 
\textbf{AUM}~\citep{pleiss2020identifying} has been described above.
\textbf{Standard} represents training with the entire dataset, without removing any samples. 
\textbf{Oracle} means training without any synthesized noisy samples.

\textbf{Experiment Setups.} 
We test our approach on 7 datasets, i.e., two synthesized noisy CIFAR datasets yet with four different noise ratios, two original CIFAR10/100 datasets, Tiny ImageNet~\citep{DBLP:conf/cvpr/DengDSLL009}, WebVision~\citep{DBLP:journals/corr/webvision-abs-1708-02862} and Clothing1M~\citep{DBLP:conf/cvpr/Clothing-XiaoXYHW15}, where the latter two are real-world noisy datasets.
Note that for a fair comparison with AUM and other baseline methods, we use WebVision50 of both Google and Flickr images and a subset of Clothing1M, named Clothing100K.
Both of them contain around 100,000 images.
For the same reason of fair comparisons with baselines, the same deep model is used for all experiments, where the model architecture is indicated for respective experiments.

\textbf{Results on Synthesized Noisy CIFAR-10/100.} 
Table~\ref{tab:synthesized datasets} displays the test accuracies on noised CIFAR-10/100, each noised by four different ratios. Our identification procedure outperforms other methods in almost every setting and surpasses oracle performance on both 20\% noisy CIFAR-10/100, indicating that some label noises in the original version of CIFAR-10/100 have been found.



\textbf{Results on Public Datasets.}
In this evaluation, the noise detector used in our approach is trained in advance on a noised CIFAR-100 and applied in an off-the-shelf manner.

Original CIFAR-10/100 and (Tiny)ImageNet datasets also contain mild label noises~\citep{DBLP:conf/nips/NorthcuttAM21}. 
We apply our method to diagnose the CIFAR-10/100 and Tiny ImageNet datasets and find a small part of images that are considered as mislabeled samples by our method.
Then we remove them from the training set and retrain the deep models.
Table~\ref{tab:real-world datasets} (three columns on the left) shows that our identification method works well on low-noise datasets and surpasses the performance of existing approaches.

WebVision50 and Clothing100K are known for their label noises and are often used to evaluate the algorithms of identifying mislabeled samples.
In Table~\ref{tab:real-world datasets} (two columns on the right), removing suspected mislabeled samples in the training set leads to a 2.8\% and 4\% enlargement of test accuracies on WebVision50 and Clothing100K, respectively. 
By use of the novel learning solution based on training dynamics, our method shows versatility in identifying mislabeled samples across model architectures, data distributions, and generations of label noises.

\subsection{Combine with Algorithm-Centric Approaches}
\label{subsec:com with SSL}
From an orthogonal aspect, we evaluate the enhancement to algorithm-centric state-of-the-art solution attributed to decontamination of training data.
Following the debugging algorithm described in \citet{mazumder2022dataperf}, we first select a number of samples with the most suspicion as label noise. 
Labels of these samples are then replaced by error-free ones, namely data debugging. 
We conduct experiments on both synthesized and real-world noisy datasets, i.e., a noised CUB-200 with 20\% symmetric noises and mini WebVision\footnote{mini WebVision denotes the Google-resourcing mini WebVision as previous works\cite{DBLP:conf/iclr/DivideMix-LiSH20,DBLP:conf/cvpr/NishiDRH21}.}.
Note that the noise detector used here is trained in advance on a noised CIFAR-100. 
We follow the same experiment setups as DivideMix and DM-AugDesc, where ResNet-50~\citep{he2016deep} and Inception-ResNet v2~\cite{DBLP:conf/aaai/SzegedyIVA17} are used respectively.

\textbf{Data Debugging Boosts.} 
Table~\ref{tab:improve on SSL} shows the performance after debugging some training instances detected by our approach. By simply applying our data-centric solution to algorithm-centric states of the art, i.e., DivideMix and DM-AugDesc, testing accuracies on both synthesized and real-world noisy datasets are further improved without any changes in model architectures or optimization processes.

\begin{table}[t]
    \caption{Testing accuracy on noisy CUB-200-2011 and original mini WebVision. D.D. denotes data debugging for short. Competitive results (better than baselines) are \textbf{bolded}.
    }
	\label{tab:improve on SSL}
	\centering
    \scalebox{0.85}{
    \begin{tabular}{ccccccc}

    \toprule
     & \multicolumn{2}{c}{CUB-200-2011 20\% Sym.} & \multicolumn{2}{c}{mini WebVision}  \\ 
    Algorithm                     & Best           & Last           & Top1           & Top5 \\ 
    
    \midrule
    \midrule
    \begin{tabular}[c]{@{}c@{}}DM-AugDesc \end{tabular}    & 79.65                & 79.01               & 78.64            & 93.20           \\
    \begin{tabular}[c]{@{}c@{}} + 5\% D.D.[Ours]\end{tabular}  & 79.55                & \textbf{79.15}      & \textbf{79.32}   & 92.56           \\
    \begin{tabular}[c]{@{}c@{}} + 10\% D.D.[Ours]\end{tabular} & \textbf{80.14}       & \textbf{79.65}      & \textbf{78.84}   & 92.84           \\ 
    \hdashline\rule{0pt}{1\normalbaselineskip}
    \begin{tabular}[c]{@{}c@{}}DivideMix \end{tabular}     & 75.96                & 73.89               & 77.32            & 91.64           \\
    \begin{tabular}[c]{@{}c@{}} + 5\% D.D.[Ours]\end{tabular}   & \textbf{76.06}       & \textbf{74.44}      & \textbf{77.84}   & \textbf{92.16}  \\
    \begin{tabular}[c]{@{}c@{}} + 10\% D.D.[Ours]\end{tabular}  & \textbf{76.77}       & \textbf{74.99}      & \textbf{78.24}   & \textbf{92.72}  \\ 
    \bottomrule
    \end{tabular}
    }
\vspace{-1em}
\end{table}

\subsection{Analyses}
\label{subsec:analyses}
We provide three analyses on the proposed noise detector.

\textbf{Tolerance to Underlying Noise. }
Underlying noises may exist in real-world datasets, even in CIFAR-10. 
The supervision information used in our approach would be affected by the unknown underlying noises.
To mimic this situation, we define a \textbf{twice-contaminated} operation.
That is to say, we contaminate the dataset twice, where the first contamination is representative of underlying noises and the second one for synthesized noises.
Empirically, we demonstrate that the noise detector can tolerate large underlying noises ratios. 

\begin{figure}[t]
\centering
\includegraphics[width=0.95\columnwidth]{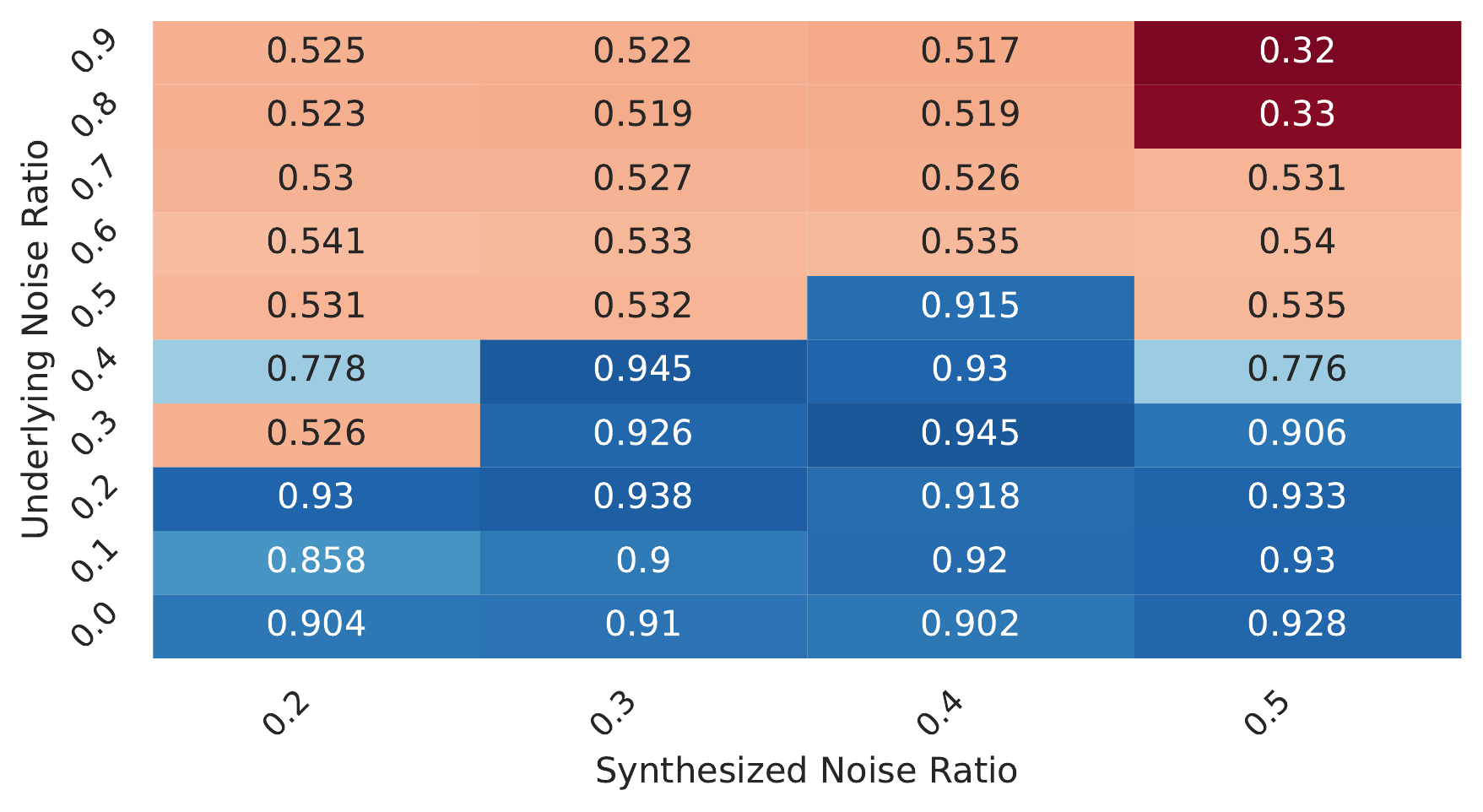}
\caption{Robustness against underlying noise in mAP. The X-axis and Y-axis represent synthesized and underlying noise ratio of a twice-contaminated CIFAR-10.}
\label{fig:map robustness}
\vspace{-1em}
\end{figure}

Figure~\ref{fig:map robustness} displays the mAP scores against the twice-contaminated datasets with underlying noises (y-axis) and synthesized noises (x-axis).
The mAP scores are measured on an independent synthesized noised Caltech256 with 40\% symmetric noises, using an LSTM network as the noise detector trained on the twice-contaminated CIFAR-10 datasets. 
Two facts are observed: 1) Under the same synthesized noise ratios, more underlying noises would harm the detector. 
2) The performance of the noise detector is satisfied if underlying noise ratio is no larger than 20\%, showing good robustness against underlying noise.

\begin{figure}[t]
\centering
\includegraphics[width=0.95\columnwidth]{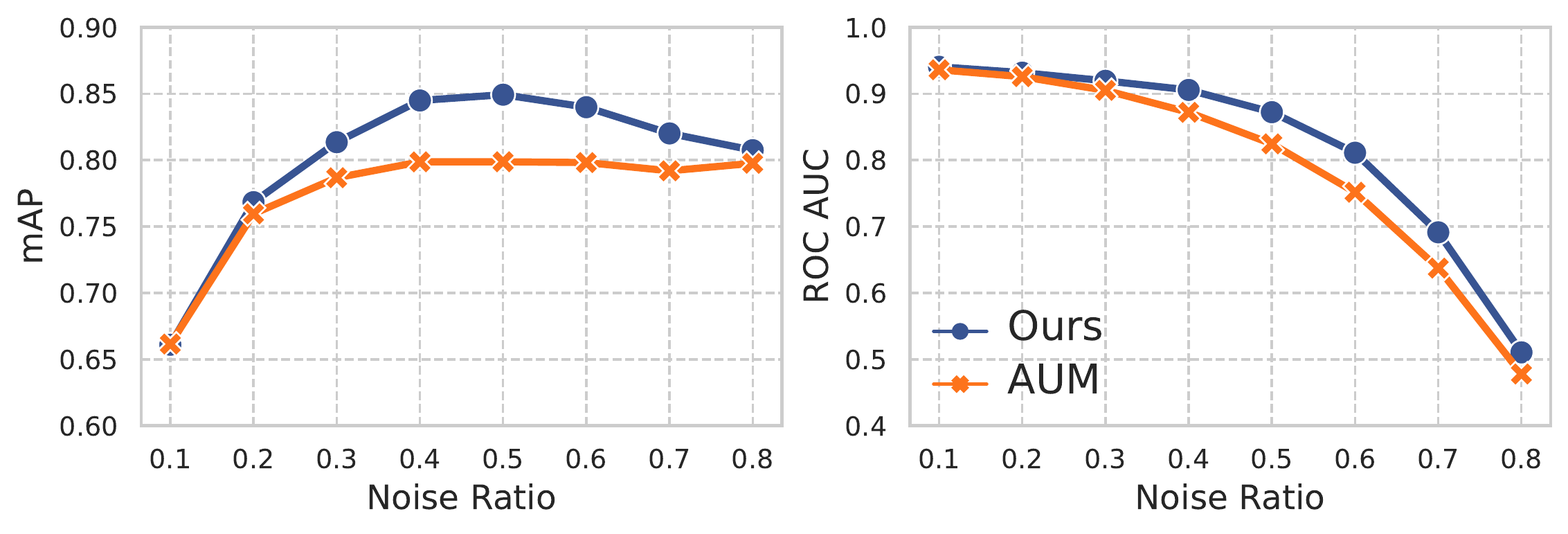}
\caption{Scores of mAP for identifying noises on CIFAR-100 where the noises are generated within super-classes.}
\vspace{-1em}
\label{fig:coarse_auc_map}
\end{figure}

\textbf{Noises within Super-Classes of CIFAR-100.}
The 100 classes in the CIFAR-100 are grouped into 20 super-classes. 
To simulate the mislabeling in the annotation process, we conduct experiments generating noises within the super-class.
This is more challenging but closer to practical scenarios.
We also report the ROC AUC and mAP scores and compare them with AUM.
Figure~\ref{fig:coarse_auc_map} demonstrates the advantage of our approach in identifying label noises within super-class in all settings of noise ratio.

\begin{figure}[t]
\centering
\includegraphics[width=0.95\columnwidth]{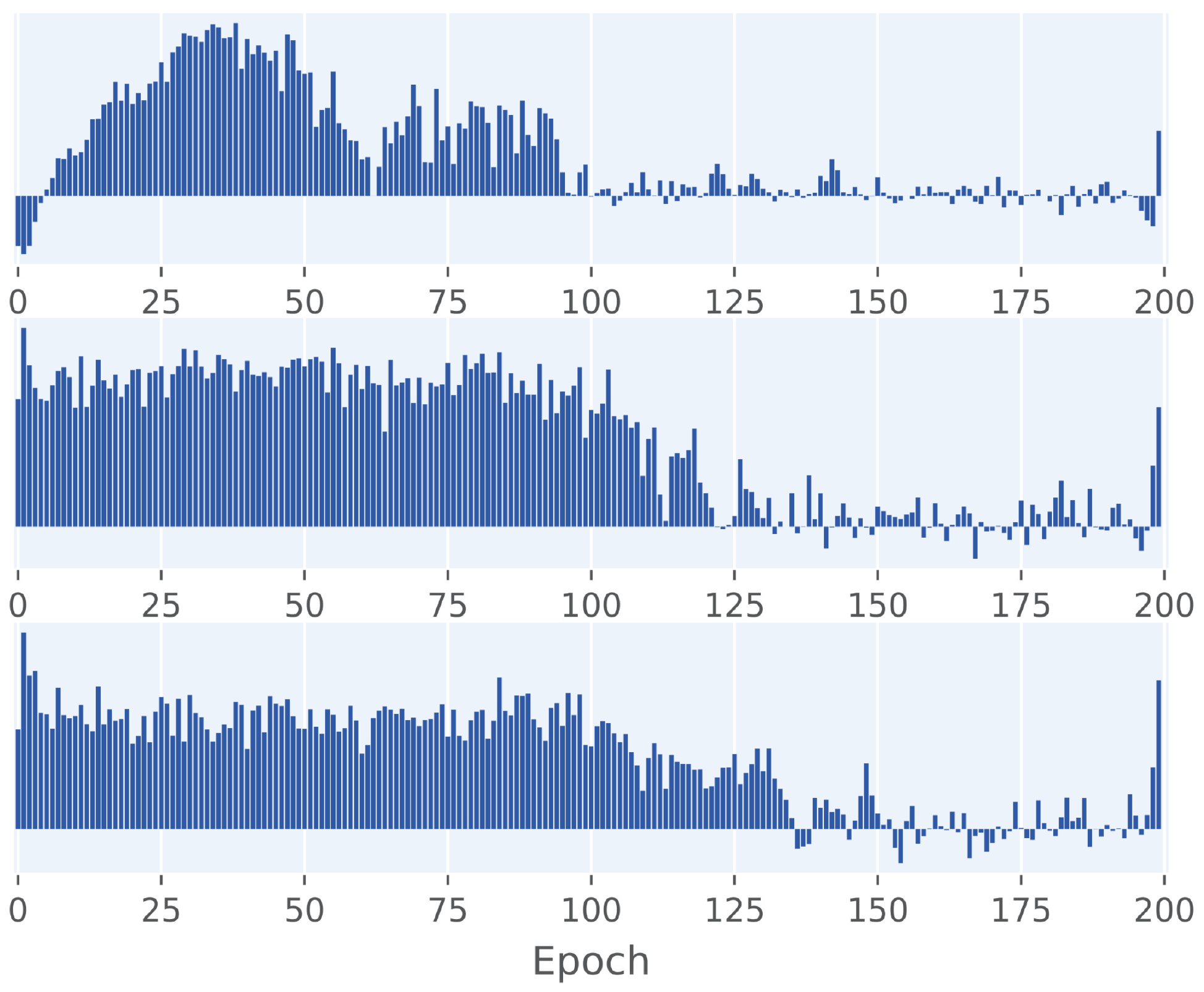}
\caption{Instance-wise feature importance given by LIME. The X-axis and Y-axis are respectively the feature index, i.e., epoch, and the importance score.}
\vspace{-1em}
\label{fig:lstm_lime}
\end{figure}

\textbf{Feature Importance Analysis through LIME.}
Though the LSTM network does not necessarily have a set of interpretable rules to identify the mislabeled samples, we could analyze the importance of features through local explanation methods, such as LIME~\citep{DBLP:conf/kdd/LIME-Ribeiro0G16}. 
Figure~\ref{fig:lstm_lime} shows the results of LIME on three mislabeled samples from noised CIFAR-10 that are correctly found by the noise detector.
From the LIME results, the first half of training dynamics (before the learning rate drops) are more important than the second half, which is mainly in accordance with AUM.
Moreover, the trained LSTM network also takes another half into consideration, where the given-label probabilities of mislabeled samples have a more significant variance than clean samples, as shown in Figure~\ref{fig:illustration}.
While the exact formula of identifying the noise is hard to deduce, the local explanations indicate that such LSTM has learned relevant features to find mislabeled samples.

\section{Conclusion}

In this paper, we introduced a learning-based approach to identify mislabeled samples from training dynamics, which trains label \emph{noise detectors} instanced by an LSTM network using synthesized noises.
Beyond handcraft features, detectors can automatically learn relevant features from the time-series data to identify the mislabeled samples in the training set. 
This trained noise detector is then applicable to other noisy datasets without further adaptation or retraining.
To validate the effectiveness of the proposed approach in detecting label noises, we conducted experiments on synthesized and real-world noisy datasets, where our approach excels the existing data-centric methods in both settings.
We further combined our approach with the mainstream algorithms of learning with noisy labels, and provided orthogonal improvements from the data aspect.
More analyses on the robustness and applicability of our approach were also provided in details.




For both researchers and practitioners, identifying mislabeled data can prelude the model development pipeline, at the sacrifice of only one additional step to obtain the corresponding training dynamics. 
Samples predicted to be mislabeled can be further re-used in other manners, such as advanced data debugging and semi-supervised learning.

\section*{Acknowledgements}
This work is supported in part by National Key R\&D Programs of China under the grant No. 2021ZD0110303.

\bibliography{aaai23}


\clearpage
\appendix

\setcounter{equation}{0}
\setcounter{subsection}{0}
\renewcommand{\theequation}{A.\arabic{equation}}
\renewcommand{\thesubsection}{\Alph{section}.\arabic{subsection}}

\section{Definitions of Evaluation Metrics}
\label{sec: metric}

There are three metrics that are used in this work, whose definitions are given in the following.
Note that the first two (Average Precision and ROC AUC) measures the alignment between predictions and ground-truths in a general view, while Precision@95 considers at a fix threshold when the recall rate reaches 95\%.

\paragraph{Average Precision}
Average Precision (AP)\footnote{\url{https://scikit-learn.org/stable/modules/generated/sklearn.metrics.average_precision_score.html}} computes the area under the precision-recall curve:
\begin{equation}
    AP = \sum_n (R_n - R_{n-1}) P_n,
\end{equation}
where $R_n$ and $P_n$ are the recall and precision values at the $n^{th}$ threshold.

\paragraph{Area under the Receiver Operating Characteristic Curve}
A receiver operating characteristic (ROC)\footnote{\url{https://scikit-learn.org/stable/modules/generated/sklearn.metrics.roc_auc_score.html}} curve, whose y-axis and x-axis denote true positive rate (TPR) and the false positive rate (FPR), respectively.
The ROC AUC computes the area under the ROC curve:
\begin{equation}
    AUC_{ROC} = \sum_n (FPR_n - FPR_{n-1}) TPR_n,
\end{equation}
where $TPR = \frac{TP}{TP+FN}$ and $FPR = \frac{FP}{FP+TN}$.

\paragraph{Precision@95}
Precision@95 entails precision value when recall equals 95\%, which means whether the detector can correctly find almost all mislabeled samples guaranteed by a recall value greater than 95\%. That is convenient for us to effectively judge the performance of noise identification.
\begin{equation}
Precision@95 = P_n, \text{ when } R_n = 0.95,
\end{equation}

\section{Complete Experiment Setups and Results}

This section introduces the additional details about the experiment setups and the supplementary experiment results.
Three evaluation experiments have been conducted in our work, and presented by Section~\ref{subsec:identify}, \ref{subsec:real-world} and \ref{subsec:com with SSL}.
The following subsections complete the main text with the same order.
Details for Analyses (Section~\ref{subsec:analyses}) are also included.

\subsection{Identify Mislabeled Samples}
In general, experiments in this part are consistent with the AUM experiment setups~\citep{pleiss2020identifying}, and all results recorded are given by the model from the last training epoch except specified.
Tables depict the mean and confidence interval given from four trials. 
Only one trial is performed on large datasets, including WebVision and Clothing100K.


\paragraph{Datasets}
For identifying the mislabeled samples, we intentionally change the labels of certain samples in the original dataset and record the mislabel information for the supervised learning of noise detectors and the evaluations.
Here, four datasets are used: CIFAR-10/100~\citep{krizhevsky2009cifar}, CUB-200-2011~\cite{wah2011caltech} and Caltech256~\cite{griffin2007caltech}.
The CIFAR-10/100~\cite{krizhevsky2009cifar} datasets separately consist of 60,000 color images of 10/100 classes with 6,000/600 images per class;
The CUB-200-2011~\cite{wah2011caltech} is a fine-grained dataset, including 11,788 images of 200 bird species;
Images from the Caltech 256~\cite{griffin2007caltech} are collected and selected from the Google Image dataset, then divided into 256 categories, with more than 80 pictures in each category.


\paragraph{Noise Generation} Datasets mentioned above are manually corrupted with symmetric and asymmetric label noises. 
Following the previous works~\citep{pleiss2020identifying,DBLP:journals/jair/ConfidentLearning-NorthcuttJC21}, symmetric noisy labels are given labels uniformly at random. 
For asymmetric ones, we alter a sample's assigned label from ground-truth class to adjacent class. 
Without further notifications, label noises are generated in such ways for all experiments.
Moreover, noises within super-class of CIFAR-100\footnote{\url{https://www.cs.toronto.edu/~kriz/cifar.html}} actually means symmetric noise among 5 like-minded classes as shown in Table~\ref{tab:coarse cifar100}.

\paragraph{Evaluation Metrics} To validate the effectiveness and robustness of our identification method at different noise ratios, we treat noise detection as a binary classification tack. We can directly measure the \textbf{ROC AUC}, \textbf{mAP} scores and \textbf{Precision@95}, mentioned above in Section~\ref{sec: metric}.
The first two measure the general performance of predicted results, while the last indicates the performance when querying most mislabeled samples.
Usually, the three metrics are positively proportional, but we report them all for completeness.

\paragraph{Models and Acquisition of Training Dynamics}
Training dynamics are computed by a conventional deep learning process with a wide range of model and training round choices. Most models can generally fit the noisy samples on the dataset ideally within the desired number of rounds, sufficient to obtain qualified training dynamics.
Precisely, we train the small version of ResNet-32~\citep{he2016deep} for CIFAR-10/100, the bottleneck version of ResNet-34 with pretrained parameters for CUB-200-2011 and ResNet-50 for WebVision50 and Clothing100K for 200 epochs, with $10^{-4}$ weight decay, SGD with the momentum of 0.9, a learning rate of 0.1, a batch size of 128. 
The learning rate becomes one-tenth at 100 and 150 epochs.
Without further notifications, the hyperparameters for getting the training dynamics are the same for all experiments.


\paragraph{Noise Detector Training}
Due to a lightweight detector model architecture, ten epochs with AdamW~\citep{DBLP:conf/iclr/AdamW-LoshchilovH19} are sufficient to reach convergence even when start from scratch, with a learning rate of 0.1 and a batch size of 64. 
For fine-tuning based on well-trained detectors migrated by noisy CIFAR-10/100 training, we recommend a learning rate of 0.03 in practice.
Without further notifications, the noise detector used in all experiments is trained on CIFAR-100 with 30\% symmetric noises without fine-tuning neither other adaptation.


\begin{figure}[h]
\centering
\includegraphics[width=0.95\columnwidth]{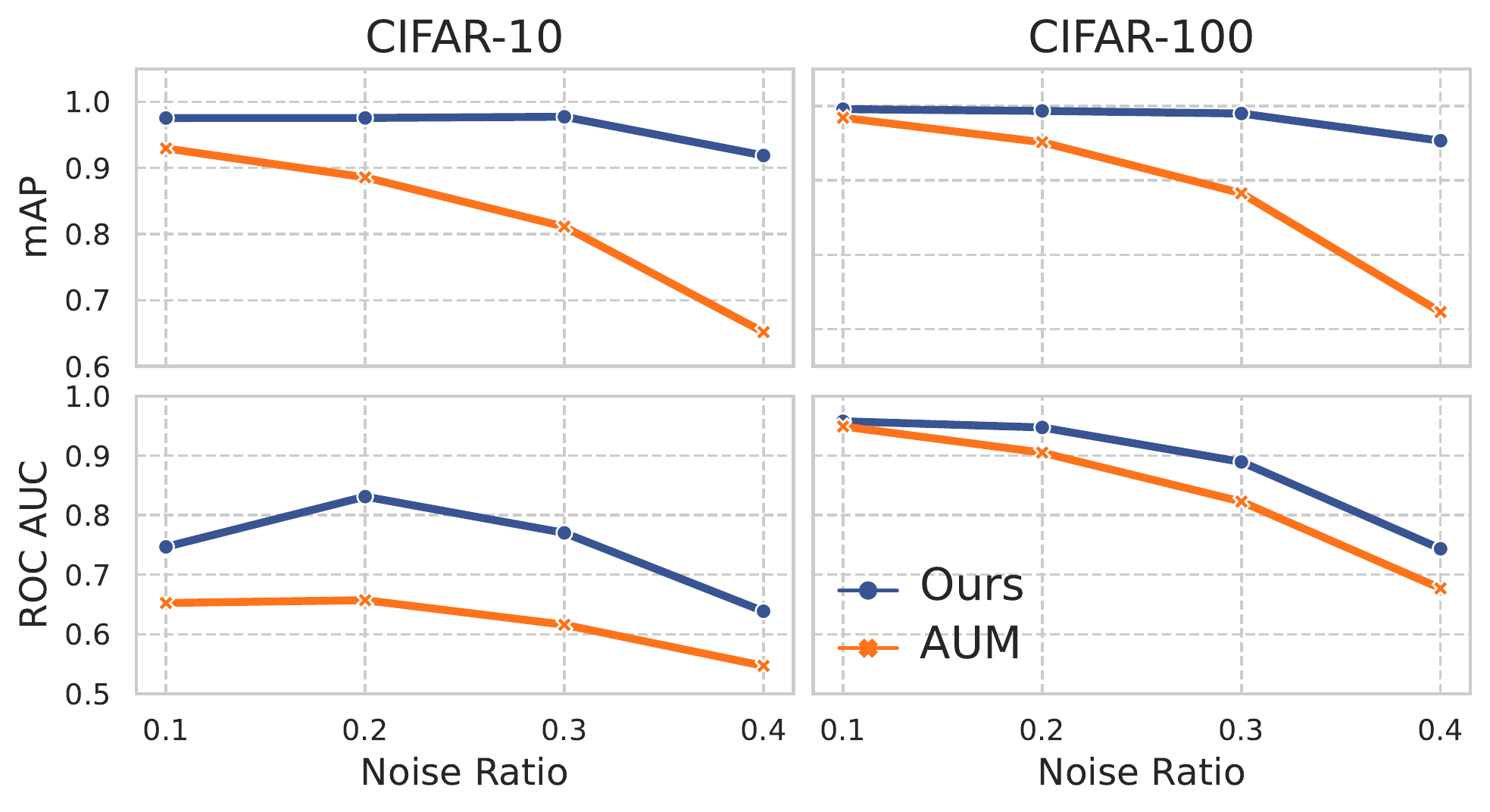}
\caption{Scores of mAP, ROC AUC and Precision@95 for identifying mislabeled samples on CIFAR-10/100 with asymmetric label noises.}
\label{fig:asym}
\vspace{-1em}
\end{figure}

\paragraph{Experiments of Asymmetric Noises}
Figure~\ref{fig:asym} depicts the mAP and AUC scores for identifying mislabeled samples in the asymmetric noises scenario on CIFAR-10/100.
Compared with symmetric noise, asymmetric noise can cause a shift in the overall data distribution, leading to a degradation in identifying mislabeled samples.
More attention should be paid to different noise distributions in future work.

\paragraph{Fine-tuning on CUB-200-2011 and Caltech-256}
Unlike hand-designed statistical features, parameters of noise detectors built on LSTM architecture can be learned, further updated and optimized when we apply the appropriate fine-tuning technology to adapt the detector on new datasets, which is the advantage of deep-learning-based methods.
In the main text, we have presented the fine-tuning results measured in mAP.
Here results for all the three metrics with/without fine-tuning leverage showed in Table~\ref{tab:migrated versus fine-tuning}.

\subsection{Retrain after Excluding Mislabeled Samples}

To validate the effectiveness of identification, we remove samples predicted to be mislabeled using the proposed approach and then retrain the deep model. 
Based on the testing accuracy after excluding samples, we are able to measure the effectiveness of mislabeled sample identification.
Within this setting, we compare our method against several existing ones, introduced in the following paragraph.

\paragraph{Datasets}
We test our approach on 7 datasets, i.e., two synthesized noisy CIFAR datasets yet with four different noise ratios, two original CIFAR10/100 datasets, Tiny ImageNet~\citep{DBLP:conf/cvpr/DengDSLL009}, WebVision~\citep{DBLP:journals/corr/webvision-abs-1708-02862} and Clothing1M~\citep{DBLP:conf/cvpr/Clothing-XiaoXYHW15}, where the latter two are real-world noisy datasets.
CIFAR datasets are both used with synthesized label noises and in their original versions.
Tiny ImageNet contains a subset of images from ImageNet rescaled to 64$\times$64.
Note that for a fair comparison with AUM and other baseline methods in this subsection, we use WebVision50 of both Google and Flickr images and a subset of Clothing1M, named Clothing100K.
Both of them contain around 100,000 images.

\paragraph{Setting of Retraining Optimization Process}
Note that the optimization process for acquisition of training dynamics is not definitely the same as the retraining process.
Here we show the settings for the retraining process.
Following the prior work setting~\citep{pleiss2020identifying}, we use SGD with Nesterov momentum, initial learning rate 0.1, a batch size of 256 for these seven datasets. 
Other settings are listed in Table~\ref{tab:experiment setting}. 
Besides, we let the batch size shrink proportionally after discarding suspected samples as the training dataset shrinks.
All other hyperparameters are the same as the original training procedure of \cite{he2016deep}.

\begin{figure}[t]
\centering
\includegraphics[width=0.95\columnwidth]{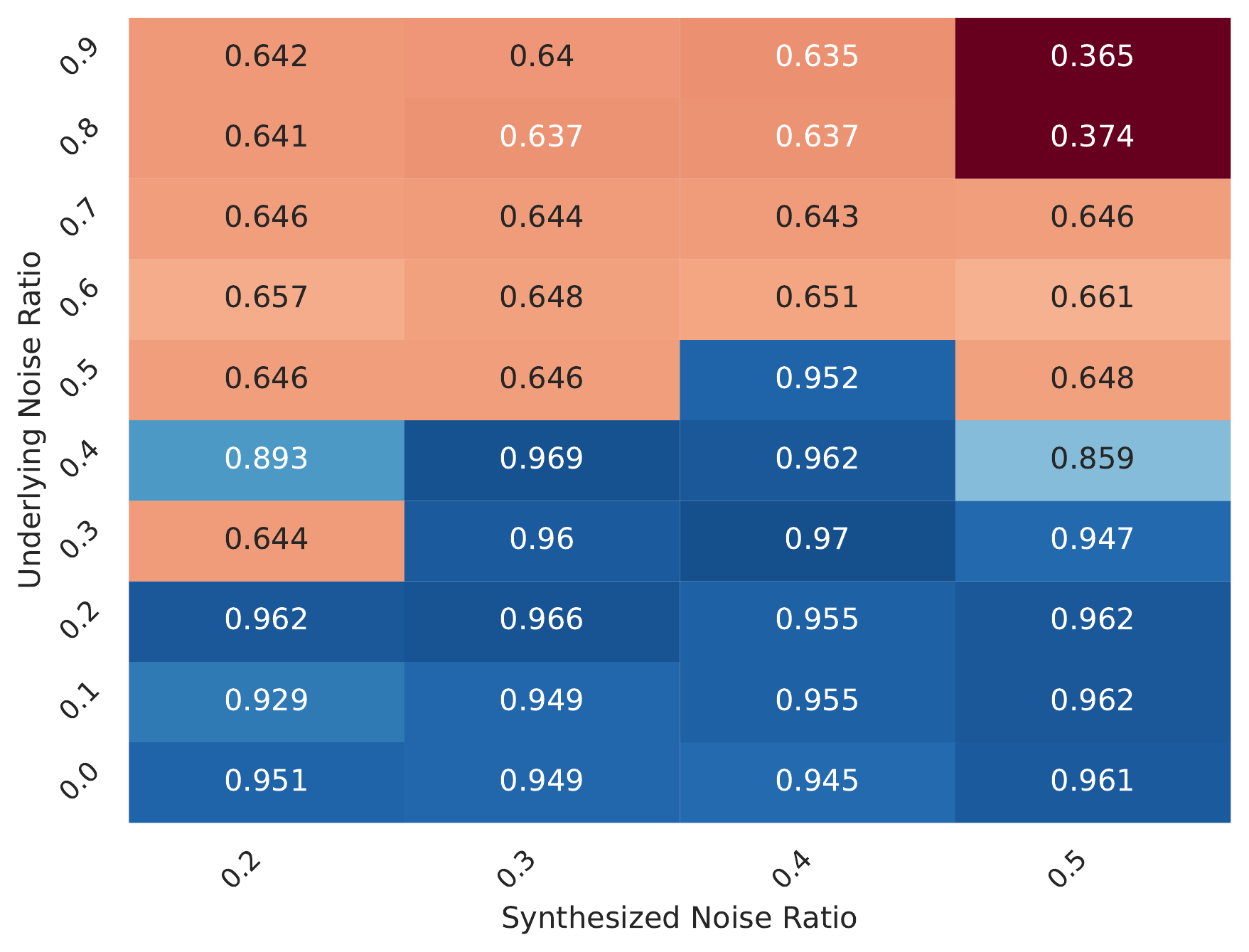}
\caption{Robustness against underlying noise in ROC AUC. The X-axis and Y-axis represent synthesized and underlying noise ratio of a twice-contaminated CIFAR-10.}
\label{fig:auc robustness}
\vspace{-1em}
\end{figure}

\begin{table*}[t]
\caption{Setting of model retraining.}
\label{tab:experiment setting}
\centering
\scalebox{0.8}{
\begin{tabular}{cccc}
\toprule
Datasets    & CIFAR, Tiny ImageNet    & CUB-200-2011, Caltech256    & mni WebVision, WebVision50, Clothing100K                                                        \\ 
\midrule
\midrule
Model       & ResNet-32         & \begin{tabular}[c]{@{}c@{}}ResNet-34 with ImageNet pretrained weights\end{tabular}         & ResNet-50                       \\ 
    \hdashline\rule{0pt}{1\normalbaselineskip}

\begin{tabular}[c]{@{}c@{}}Length of training \\ Milestones of learning rate drop\end{tabular} & \begin{tabular}[c]{@{}c@{}}300 epochs\\ 150 and 225\end{tabular} & \begin{tabular}[c]{@{}c@{}}300 epochs\\ 150 and 225\end{tabular}         & \begin{tabular}[c]{@{}c@{}}180 epochs\\ 60 and 120\end{tabular}          \\ 
    \hdashline\rule{0pt}{1\normalbaselineskip}
Data augmentation                                                                         & \begin{tabular}[c]{@{}c@{}}Random horizontal flips\\ Random crops\end{tabular}          & \begin{tabular}[c]{@{}c@{}}Random horizontal flips\\ Random crops\\ Random scaling\end{tabular} & \begin{tabular}[c]{@{}c@{}}Random horizontal flips\\ Random crops\\ Random scaling\end{tabular} \\ 
\bottomrule
\end{tabular}
}
\end{table*}

\subsection{Combine with Algorithm-Centric Approaches}

From an orthogonal aspect, we evaluate the enhancement to algorithm-centric state-of-the-art solution attributed to decontamination of training data.

\paragraph{Datasets}
We conduct experiments on both synthesized and real-world noisy datasets, i.e., a noised CUB-200 with 20\% symmetric noises and mini WebVision.
This mini version of WebVision is different from the one used in the previous subsection.
The mini WebVision denotes the Google-resourcing mini WebVision as previous works~\cite{DBLP:conf/iclr/DivideMix-LiSH20,DBLP:conf/cvpr/NishiDRH21}.

\paragraph{Baselines and Models}
We follow the same experiment setups as DivideMix and DM-AugDesc, where ResNet-50~\citep{he2016deep} and Inception-ResNet v2~\cite{DBLP:conf/aaai/SzegedyIVA17} are used respectively.

\paragraph{Data Debugging}
Without any prior knowledge on the label noises on the dataset, choices of small percentage for which the labels are manually verified, or automatically yet imperfectly corrected by the model trained by excluding a large part of suspicious samples.
In our experiments, we have evaluated two choices: 5\% and 10\%, both yielding improvements over the algorithm-centric methods alone, as shown in Table~\ref{tab:improve on SSL} in the main text.
Specifically, on noised CUB-200, a pseudo human verification is performed for the available ground-truth.
While on mini Webvision, we first predict and exclude the mislabeled samples using the trained noise detector, then train a model (with the same architecture) without using the algorithm-centric method, and finally correct the top 5\% (10\% respectively) using the predictions of the trained model.
As we can see, by applying our data-centric solution to algorithm-centric states of the art, i.e., DivideMix and DM-AugDesc, testing accuracy on both synthesized and real-world noisy datasets are further improved without any changes in model architectures or optimization processes.

\subsection{Analyses}

Details for analyses are also included.

\paragraph{Tolerance to Underlying Noise}
Underlying noises may exist in real-world datasets, even in CIFAR and ImageNet. 
The supervision information used in our approach would be affected by the unknown underlying noises.
To mimic this situation, we define a \textbf{twice-contaminated} operation.
That is to say, we contaminate the dataset twice, where the first contamination is representative of underlying noises and the second one for synthesized noises.
Empirically, we demonstrate that the noise detector can tolerate large underlying noises ratios. 

Figure~\ref{fig:map robustness} displays the mAP scores against the twice-contaminated datasets with underlying noises (y-axis) and synthesized noises (x-axis).
Figure~\ref{fig:auc robustness} represents the robustness against underlying label noises measured in ROC AUC scores, to complete the results in Figure~\ref{fig:map robustness}.

\paragraph{Noises within Super-Classes of CIFAR-100}
The 100 classes in the CIFAR-100 are grouped into 20 super-classes. 
To simulate the mislabeling in the annotation process, we conduct experiments generating noises within the super-class.
This is more challenging but closer to practical scenarios.
We have reported the ROC AUC and mAP scores and compare them with AUM in Figure~\ref{fig:coarse_auc_map}, which demonstrates the advantage of our approach in identifying label noises within super-class in all settings of noise ratio.
Here we provide the details of the super-classes in CIFAR-100.

\begin{table*}[t]
\caption{Super-classes in CIFAR-100.}
\label{tab:coarse cifar100}
\centering
\scalebox{1}{
\begin{tabular}{|l|l|}
\hline
\textbf{Superclass}           & \textbf{Classes}                                 \\ \hline
aquatic mammals               & beaver, dolphin, otter, seal, whale              \\ \hline
fish                          & aquarium fish, flatfish, ray, shark, trout       \\ \hline
flowers                       & orchids, poppies, roses, sunflowers, tulips      \\ \hline
food containers               & bottles, bowls, cans, cups, plates               \\ \hline
fruit and vegetables          & apples, mushrooms, oranges, pears, sweet peppers \\ \hline
household electrical devices   & clock, computer keyboard, lamp, telephone, television \\ \hline
household furniture           & bed, chair, couch, table, wardrobe               \\ \hline
insects                       & bee, beetle, butterfly, caterpillar, cockroach   \\ \hline
large carnivores              & bear, leopard, lion, tiger, wolf                 \\ \hline
large man-made outdoor things & bridge, castle, house, road, skyscraper          \\ \hline
large natural outdoor scenes  & cloud, forest, mountain, plain, sea              \\ \hline
large omnivores and herbivores & camel, cattle, chimpanzee, elephant, kangaroo         \\ \hline
medium-sized mammals          & fox, porcupine, possum, raccoon, skunk           \\ \hline
non-insect invertebrates      & crab, lobster, snail, spider, worm               \\ \hline
people                        & baby, boy, girl, man, woman                      \\ \hline
reptiles                      & crocodile, dinosaur, lizard, snake, turtle       \\ \hline
small mammals                 & hamster, mouse, rabbit, shrew, squirrel          \\ \hline
trees                         & maple, oak, palm, pine, willow                   \\ \hline
vehicles 1                    & bicycle, bus, motorcycle, pickup truck, train    \\ \hline
vehicles 2                    & lawn-mower, rocket, streetcar, tank, tractor     \\ \hline
\end{tabular}
}
\end{table*}

\nobibliography*


\end{document}